%% file: bmvc_paper.tex
\documentclass{bmvc2k}

\title{GeoFormer: \\ A Multi-Polygon Segmentation 
Transformer}

\addauthor{Maxim Khomiakov}{maxk@dtu.dk}{1}
\addauthor{Michael Riis Andersen}{miri@dtu.dk}{1}
\addauthor{Jes Frellsen}{jefr@dtu.dk}{1}

\addinstitution{
 Department of Applied Mathematics and Computer Science\\
 Technical University of Denmark\\
 Kgs.\@ Lyngby, Denmark
}

\runninghead{GeoFormer}{A Multi-Polygon Segmentation Transformer}

\begin{document}

\maketitle

\begin{abstract}
    In remote sensing, there exists a common need for learning scale invariant shapes of objects like buildings. Prior works rely on tweaking multiple loss functions to convert segmentation maps into the final vectorised representation, necessitating arduous design and optimisation. For this purpose, we introduce the GeoFormer, a novel architecture that presents a remedy to the said challenges, learning to generate multi-polygons end-to-end. By modelling keypoints as spatially dependent tokens in an auto-regressive manner, the GeoFormer outperforms existing works in delineating building objects from satellite imagery. We evaluate the robustness of the GeoFormer against former methods through a variety of parameter ablations and highlight the advantages of optimising a single likelihood function. Our study presents the first successful application of auto-regressive transformer models for multi-polygon predictions in remote sensing, suggesting a promising methodological alternative for building vectorisation.
\end{abstract}

\input{sec/1_intro}
\input{sec/2_background}
\input{sec/3_methods}

\input{sec/4_experimental_setup}

\input{sec/5_results}

\input{sec/6_ablations_robustness}

\input{sec/7_discussion}
\input{sec/8_conclusion}
\newpage
\appendix
\input{sec/X_suppl}

\bibliography{egbib}
\end{document}

%% file: sec/1_intro.tex
\begin{figure*}[htb]
  \centering
  \includegraphics[width=1\textwidth]{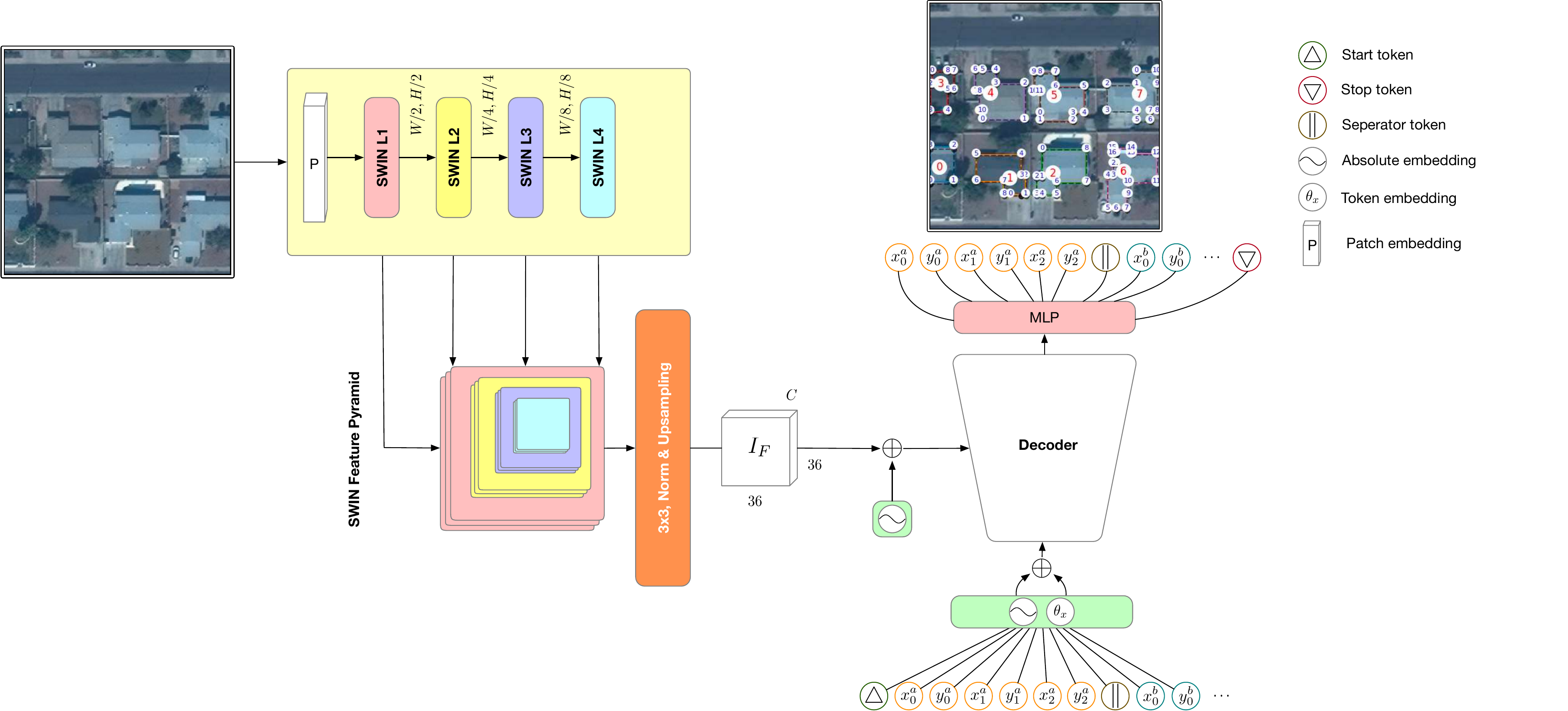}
  \caption{
  Illustration of the GeoFormer model architecture. On the left hand side an image is patch embedded and parsed through all four SWINv2 \cite{liu2022swin} layers, whilst each layer skips forward the feature representation which is then convolved and upsampled to match the hidden dimensions of the decoder. On the right hand side our auto-regressive decoder, which takes as input a flattened sequence of spatial tokens together with 3 special tokens.
  }
  \label{fig:model_illustration}  
\end{figure*}

\section{Introduction}
\label{sec:intro}

In recent years, we have witnessed a convergence in machine learning (ML), with cross-pollination of methods from vision and natural language processing (NLP) yielding impressive results in new contexts. The adaptation of the NLP-originated Transformer model \cite{vaswani2017attention} in vision research has become increasingly prevalent, with researchers discovering the benefits of applying multi-headed attention modules to domain-specific problems. Recent studies demonstrate the unification capabilities of these models \cite{chen2022unified,you2023ferret,zou2023generalized}, where a single model, conditioned on different contexts, can perform diverse tasks.

This paper proposes a novel architecture and demonstrates the utility of image-to-sequence auto-regressive transformer models in the challenging task of vectorising building objects from satellite imagery. To the best of our knowledge, this is the first successful demonstration of such a model learning to generate a sequence of multiple building polygons in a remote sensing context, relying solely on a single likelihood function. It is also the first instance of a deep generative model trained end-to-end from scratch outperforming former methods on the popular Aicrowd Mapping Challenge \cite{mohanty2020deep} building delineation benchmark dataset. Unlike comparable studies \cite{Girard2020,Li2019,xu2023hisup,Alidoost2019,zorzi2022polyworld,liu2023polyformer}, our model does not require hyperparameter tuning of individual loss term weights, raster-to-vector post-processing methods, or exogenous predictions, such as bounding box (pre)-training \cite{liu2023polyformer}, which are easily inferred if predicting correctly identified keypoints.

The challenge of generating ordered keypoints to form a polygon shares similarities with image captioning tasks, where the goal is to predict an ordered sequence corresponding to the context of an input image. This inspired the use of image-to-sequence models, as demonstrated in previous image captioning works \cite{vinyals2015show,xu2015show}.

Another crucial motivation is addressing the sparsity in vectorised polygons, learning the minimal set of keypoints mapping to geometric properties of objects (e.g., corners or intersections between linear surfaces). Vector graphics representations, being scale invariant and consisting of points and edges, are highly applicable in remote sensing, especially for buildings \cite{girard2021polygonal}. However, most comparable methods rely on semantic segmentation when predicting building polygons \cite{xu2023hisup,zorzi2022polyworld,Alidoost2019,li2021joint,li2023joint-followup,Bittner2018}, optimising for criteria not wholly aligned with the task at hand. This approach often leads to high-confidence points within the inner hull of the building polygon, with the highest uncertainty around the edges, while the inner hull points are rarely of interest. This can prove suboptimal for downstream applications \cite{marcos2018learning}, as right-angled geometric features are not immediately apparent, leading to precision errors when applied in canonical mapping or geospatial applications. Traditionally, researchers have employed heuristic or learned approaches \cite{Zorzi2020} to convert segmentation maps into accurate vectorised polygons.  
In summary, our paper proposes an alternative approach that not only matches but exceeds the performance of prior works by a large margin while producing a directly usable output representation in vector format end-to-end.

Our contributions are as follows: 

\begin{itemize}
    \item We introduce GeoFormer\footnote{Code available at: \url{https://github.com/pihalf/GeoFormer}}, a novel encoder-decoder architecture with a pyramidal image feature map that learns spatially dependent tokens sequentially.
    \item GeoFormer presents a substantial improvement over state-of-the-art performance on a popular benchmark dataset
    \item Finally, we study the robustness of our model relative to previous methods and present an ablation study to motivate our model design choices.
\end{itemize}

%% file: sec/2_background.tex
\section{Previous works}
In the domain of building delineation and vectorised polygon prediction, studies typically fall into one of three categories: semantic segmentation, regression, or auto-regressive-based approaches. Most prior state-of-the-art methods, and those achieving the best performance on our benchmark dataset, begin with a U-Net-based architecture \cite{ronneberger2015u}, utilising either rule-based or learned methods to process raster output into vector shapes. In regression-based studies, the general approach involves producing bounding box predictions alongside an output representation with coordinates for each sample. The output sequence is fixed beforehand and relies on learned or rule-based filtering for the final result. Auto-regressive methods, on the other hand, treat output coordinates as a dynamically sequential prediction task, forming an image representation from which a learned model attempts to generate a sequence of points contingent on accurately predicting prior points.

\paragraph{Semantic Segmentation Based Approaches}
Semantic segmentation is widely used for building delineation tasks \cite{zhao2021building,hu2022polybuilding}, with neural networks learning to distinguish buildings from the background. Subsequent vectorisation procedures might include techniques ranging from the Douglas-Peucker algorithm to more sophisticated learned approaches \cite{zorzi2022polyworld,girard2021polygonal,xu2023hisup}. While semantic segmentation is a natural fit for this task, it has limitations, particularly in converting segmented model outputs into vectorised forms with hard edges \cite{marcos2018learning}. The issues primarily stem from the pixel-based optimisation of models, which is not ideal for sparse vectorised representations due to scale variance and a tendency for models to show the highest uncertainty around building edges. This necessitates further post-processing, whether heuristic or learned, to produce usable results for downstream tasks. Several recent state-of-the-art methods follow this approach, using U-Net-style architectures as a starting point \cite{xu2023hisup,zorzi2022polyworld,girard2021polygonal,zhang2023hit,li2021joint,li2023joint-followup}. Our study proposes an alternative approach that could offer better results depending on the use case.

\paragraph{Regression Based Approaches}
Recent studies have explored building vectorisation through regression on output coordinates \cite{hu2022polybuilding,chen2022heat,zhang2023hit}. These studies utilise encoder-decoder transformer architectures, with \citet{hu2022polybuilding} adapting Deformable-DETR \cite{zhu2020deformable} to predict coordinates, bounding boxes, and point classifications consecutively. \citet{zhang2023hit} employ a Region Proposal Network for identifying regions of interest and a Transformer Encoder-Decoder for the final polygonal representation. HEAT \cite{chen2022heat} addresses a related problem, predicting multi-polygons of buildings using multiple models: initially identifying building corner candidates, followed by two decoders to filter and connect these corners.

\paragraph{Auto-Regressive Based Approaches}
Inspired by developments in NLP, recent studies have adapted language-based models for vectorisation tasks, showing significant promise in image object vectorisation and related tasks. Image captioning models, for instance, have been formulated as auto-regressive categorical distribution prediction tasks \cite{vinyals2015show,xu2015show}, leading to advancements in models like Polygon-RNN, Polygon-RNN++, and PolyMapper \cite{Urtasun2017,Acuna2018,Li2019,zhao2021building}. These models share a sequence-to-sequence framework, predicting vertices auto-regressively. The models rely on a probability distribution defined as $p(x_t|I,x_{0},x_{t-2},x_{t-1})$, where  $x_t$ is the current token conditioned on the latent image representation $I$, as well as the initial, $x_{0}$, and two preceding tokens, $x_{t-2},x_{t-1}$. Similar approaches have been applied to learning 2D polygons \cite{khomiakov2023polygonizer} and 3D meshes \cite{polygen,khomiakov2023learning}. Recent studies, e.g., \cite{chen2022unified,liu2023polyformer}, have further demonstrated the versatility of these models, with \cite{chen2022unified} showing a common architecture learning multiple tasks and PolyFormer \cite{liu2023polyformer} achieving state-of-the-art performance in text-guided polygonisation tasks.

\paragraph{The GeoFormer}
Building upon prior work, the GeoFormer presents a novel alternative for which there has not been a successful result so far. Unlike earlier studies limited to predicting a single object per scene \cite{Urtasun2017,Acuna2018,khomiakov2023polygonizer}, the GeoFormer enables multi-object polygon detection in a remote sensing context without reliance on exogenous data (e.g., bounding boxes) or pre-training \cite{liu2023polyformer}. Our aim for this study is to demonstrate the GeoFormer's performance, generalisation, and robustness capabilities compared to previous studies, highlighting its effectiveness, particularly for multi-object polygon detection with purely tabula rasa training, in direct comparison with raster-based methods.

%% file: sec/3_methods.tex
\section{Methods}%
Our goal is to predict the coordinates of all polygons in a scene. We discretize all coordinates and represent the set of polygons as a flattened sequence of $x,y$-coordinates. Let $x_{i}^n, y_{i}^n$ be the coordinates of the $i$'th vertex of the $n$'th polygon, then $s$ is the sequence of flattened coordinates defined by
\begin{align}
    s = \left[\textcolor{starttoken}{\bigtriangleup},\textcolor{orange}{x_{0}^1,y_{0}^1,x_{1}^1,y_{1}^1,}\ldots,\textcolor{septoken}{||}\textcolor{teal}{,x_{0}^2,y_{0}^2,}\ldots,\textcolor{stoptoken}{\bigtriangledown}\right]
    = \left[s_{0},\textcolor{orange}{s_1,s_2,s_3,s_4},\ldots,s_T \right],
\end{align}
where $\textcolor{starttoken}{\bigtriangleup},\textcolor{septoken}{||},\textcolor{stoptoken}{\bigtriangledown}$ are three special tokens representing different events in our model, namely start, separation, and stop. While start and stop tokens are commonly used, the inclusion of the separator token $||$ tells the model to proceed to the next object of interest. The model relies on a sequence-to-sequence approach leveraging an image encoder $\mathcal{F}(I)$ and a geometric-token decoder 
$\mathcal{D}(s_{<t},\mathcal{F}(I))$, which predicts the conditional probability of $p(s_{t} | I,s_{<t})$ where $t$ is the current position of the sequence. Both of our models consist of Transformer-based architectures \cite{vaswani2017attention}, and we write the likelihood of observing sequence $s$ given an image $I$ as
\begin{align}
    p_{\theta}(s|I) = \prod_{t=0}^{T} p_{\theta}(s_{t}|I,s_{<t}),
\label{eqn:liklihood}
\end{align}
where $\theta$ is the parameters of the model. The distribution is defined over the discretized coordinates space plus three special tokens, thus $s_t \in \left\lbrace 0, 1, 2,  \ldots, |I_{D}| + 3 \right\rbrace$ where $|I_{D}|$ is the width dimension of the image, which is equal to the height dimension. Thus, we seek to learn the parameters $\theta$ of the network parametrizing the conditional categorical distribution, $p_\theta$, by minimizing the negative log-likelihood, which for a single image $I$ and corresponding sequence $s$ is given by
\begin{align}
    \mathcal{L} = - \sum_{t=0}^T \log p_{\theta}(s_t|I,s_{<t}) .
    \label{eqn:loglik_loss}
\end{align}

\subsection{Image encoder}
We leverage an image encoder based on a SWINv2-backbone \cite{liu2022swin}, which allows us to learn hierarchical feature maps throughout the image through different levels of granularity with the help of window functions. As shown in Figure \ref{fig:model_illustration}, the image encoder model initially follows the classical ViT transformer \cite{dosovitskiy2020image} approach, in which the input image $I \in \mathbb{R}^{H \times W \times C}$ is split into smaller patches $I_{p} \in \mathbb{R}^{N \times (P^2 \times C)}$, where $P$ is the chosen patch size for each of the patches. Following the patch embedding, we extract feature representations from each of the four layers in SWINv2 \cite{liu2022swin}, such that we produce the feature maps of dimensions $[ H\times W \times C,H/2\times W/2 \times C,H/4\times W/4 \times C,H/8\times W/8 \times C ]$. In order to fuse these feature maps across different scales, the intermediate representations are then convolved with a 3x3 kernel with batch-norm and upsampling to produce pyramidal features maps, which gives the final feature representation $I_{F}$ as $36\times 36 \times C$. Following the extraction of our final image feature representation $I_{F}$, we apply a learned and absolute positional bias by spanning a meshgrid in 2D evaluated in a $36 \times 36$ grid with sinusoidal frequency functions representing each pixel of the feature map in $I_{F}$ with a distinct value, together with a learned parameter matrix $\theta_{p} \in \mathbb{R}^{36 \times 36}$ similar to \cite{polygen,chen2022heat}, in order to cement the spatial embeddings. The final image feature representation $I_{F}$ now serves as the keys and values in the cross-attention of our decoder.

\subsection{Geometric decoder}
For the auto-regressive decoder we start with the canonical Transformer \cite{vaswani2017attention} formulation, where each transformer block consists of a set of multi-head attention layers (MHA) being the causal multi-head attention at the input, and the cross multi-head attention (CMHA) in the middle. The input to the attention mechanism in the form of queries $\mathbf{Q}_d$, which are of the dimensions $ T \times d$, where $d$ is the embedding dimension of both the transformer encoder $\mathcal{F}(I)$ as well as the decoder $\mathcal{D}(s_{<t},I_{F})$, and $T$ is the input sequence length. The keys and the values $\mathbf{K}_d,\mathbf{V}_d$ have the dimensions $T' \times d$, where $T'$ is the number of image patches. We can formulate the decoder as a set of repeated operations: Causal multi-head attention (MHA), layer normalization (LN) \cite{ba2016layer}, and cross multi-head attention (CMHA) and Feedforward (FFN) operations . MHA takes as input $\mathbf{Q,K,V}$ denoted as queries, keys, and values which at the initial layer is the input data sequence $s$ projected onto the $d$-embedding dimensions of the Decoder. We can write the MHA block as:
\begin{align}
    \mathbf{Q} &= sW^{q}, \mathbf{K} = sW^{k}, \mathbf{V} = sW^{v} \\
    z^{(m)} &= \sigma\left( \dfrac{q^{(m)}k^{(m)^T}}{\sqrt{d}} \right) v^{m} \\
    z &= \text{Concat} (z^{1},z^{2},...,z^{M})W^o ,
\end{align}
where $m = 1,\ldots,M$ and $s \in \mathbb{R}^{T \times d}$, and each of a total M attention heads has the dimensions $d_{\text{head}} = d / M$, $\mathbf{Q}^{(m)},\mathbf{K}^{(m)},\mathbf{V}^{(m)} \in \mathbb{R}^{T \times d}$ $W^q,W^k,W^v,W^o \in \mathbb{R}^{d\times d}$ and $\sigma$ is the softmax function. For causal multi-head attention, we add an upper triangular mask with negative infinity values, which masks out all values inside the softmax function of the scaled dot-product $\mathbf{Q}^{(m)}\mathbf{K}^{(m)^T}/\sqrt{d}$, such that only values at $s_t$ or before $s_{<t}$ can be attended to. For the cross multiheaded attention the queries are formed by the previous MHA block whilst the keys and the values are provided by the encoder. As such, we can formulate the set of transformer decoder blocks as
\begin{align}
    z'_{l} &= \text{MHA}(\text{LN}(z_{l-1})) + z_{l-1} \\
    z_L &= \text{SiLU}(\text{FFN}(\text{LN}(z_{l}^{'}))+z'_{l}),
\end{align}
where $z_l$ is the latent representation at intermediate transformer layers, FFN is the feed forward layer, and $z_L$ is the final layer in the transformer decoder, which is activated with a Sigmoid Linear Unit activation function \cite{hendrycks2016gaussian}. For the decoder with cross attention, a repeated set of $z_l$-modules are being computed but now without a causal mask and with keys and values being provided by the encoder.

Transformers leverage the self-attention operation, which is inherently permutation equivariant. Thus an important component of our model's design is how we incorporate positional information. For the decoder, we leverage a combination of positional embeddings: sinusoidal absolute positional embeddings, learned token embeddings, and relational rotary embeddings \cite{su2021roformer}. To allow the model to attend to distant tokens while emphasizing nearby tokens due to the multi-object nature of the task (where parts of the feature space should be localized per building object), we apply Added Linear Bias (ALiBi) \cite{press2021train} for a third of the attention heads in each attention block of our decoder. Additionally, we have incorporated Rotational Positional Embeddings (RoPE) \cite{su2021roformer} as part of our model to facilitate the learning of the relative spatial dependencies between tokens for our problem. Both RoPE and ALiBi has the benefit of re-enforcing the token positions during each layer, as opposed to canonical positional embeddings applied in the beginning \cite{vaswani2017attention}.

%% file: sec/4_experimental_setup.tex
\section{Experimental setup}
We used the Aicrowd Mapping Challenge dataset \cite{mohanty2020deep}, a popular benchmark for building delineation using satellite imagery, which contains 341,058 images (280,741 for training and 60,317 for validation) with an image resolution of $300 \times 300$ and a spatial resolution of 30\,cm per pixel. The dataset has 2,395,553 building instances in the training set and 515,364 annotations for the validation set. We sampled 5\% of the training dataset for validation and used the competition validation set as our test set. Data augmentation techniques, such as random rotations, horizontal and vertical flips, and color jitter, were applied during training \cite{xu2023hisup,Girard2020}.

For model training, we scaled the images and annotations to $224 \times 224$. The annotations were sorted relative to the Euclidean distance to the image centroid, clockwise around the origin, to preserve object order. We apply a patch size of 4 and a window size of 7 for the encoder with hidden dimensions of 512. Likewise the decoder is composed of 8 layers each with 24 attention heads. Further details of hyperparameters are described in the supplementary materials.

\subsection{Evaluation measures}
To measure our model's efficacy, we apply a set of benchmark dataset metrics. The main competition metrics are segmentation mask-based, following the MSCOCO object detection metrics \cite{lin2014microsoft}, including mean average precision (AP) and mean average recall (AR), measured using different IoU (intersection over union) thresholds. 

In addition to the canonical MSCOCO metrics, we include measures focusing on polygon fidelity, such as the mean average precision at the boundary of each mask \cite{cheng2021boundary}, the Complexity-aware IoU \cite{zorzi2022polyworld}, which discounts the IoU score relative to the fraction of predicted points, and the PoLiS measure \cite{polisDistance}, which is the average minimal distance between predicted vertices and ground truth. These measures (bAP, C-IoU, and PoLiS) are computed for polygons with more than 50\% overlap with the ground truth to ascertain which prediction belongs to which object. Finally, we measure the global IoU and the N-ratio, which compares the cardinality of our predicted polygon to the ground truth.

%% file: sec/5_results.tex
\begin{figure}[tb]
    \centering
    \includegraphics[width=1\textwidth]{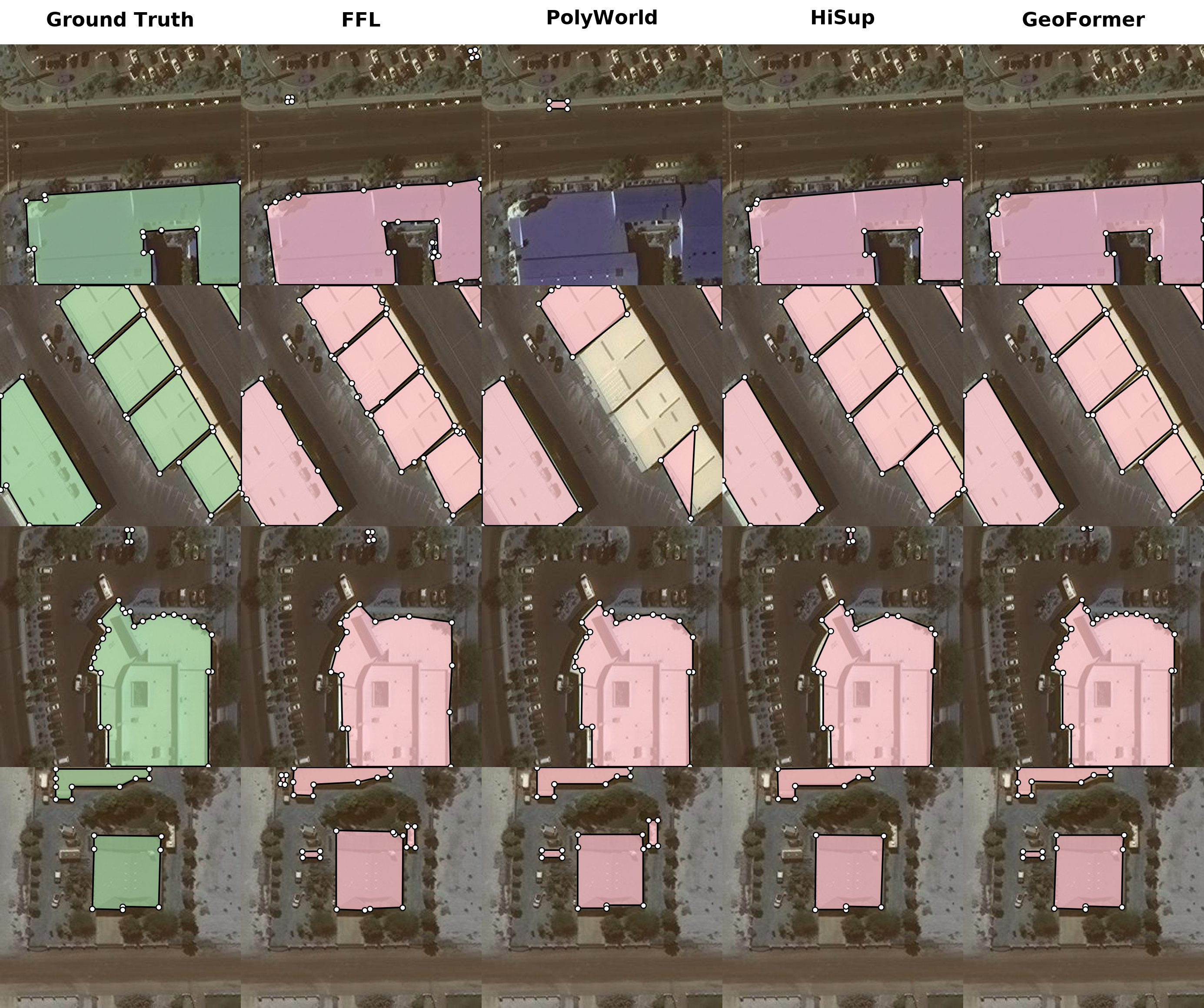}
    \caption{Qualitative results of model predictions on test set images together with ground truth. Columns from left to right: ground truth, FFL \cite{girard2021polygonal}, PolyWorld \cite{zorzi2022polyworld}, HiSup \cite{xu2023hisup} and GeoFormer (ours). }
    \label{fig:qual-results-main}
\end{figure}

\section{Results}

\begin{table}[tb]
\caption{Results on benchmark dataset for Aicrowd \cite{mohanty2020deep}. Best performance marked in bold. The values are provided by the respective paper authors, while $^\star$-values are provided by authors of HiSup \cite{xu2023hisup}.}
\vskip 0.05in
\centering
\resizebox{1\textwidth}{!}{
\begin{tabular}{lrrrrrrrrccrc}
\toprule
\textbf{Method} & \textbf{AP}$\uparrow$ &  \textbf{AP}$_{50}$$\uparrow$ &  \textbf{AP}$_{75}$$\uparrow$ &    \textbf{AR}$\uparrow$ &   \textbf{AR}$_{50}$$\uparrow$ &   \textbf{AR}$_{75}$$\uparrow$ & \textbf{bAP}$\uparrow$ &    \textbf{IoU}$\uparrow$ &   \textbf{C-IoU}$\uparrow$ &  \textbf{PoLiS}$\downarrow$  &  \textbf{N-ratio} \\
\midrule
 \text {PolyMapper$^\star$ \citep{Li2019}}& 55.7 & 86 & 65.1 & 62.1 & 88.6 & 71.4 & 22.6& 77.6& 65.3&  2.215&1.29\\
 \text {FFL$^\star$ \citep{Girard2020} }& 67.0& 92.1& 75.6& 73.2& 93.5& 81.1& 34.4& 84.3& 73.8&  1.945&1.13\\
 \text {PolyWorld \citep{zorzi2022polyworld} } & 63.3 & 88.6 & 70.5 & 75.4 & 93.5 & 83.1 & 50.0& 91.2& 88.2&  0.962&0.93\\
 \text {W. Li et al. \citep{li2021joint} } & 73.8 & 92 & 81.9 & 72.6 & 90.5 & 80.7 & -& -& -& -&-\\
 \text {PolyBuilding \citep{hu2022polybuilding} } & 78.7 & 96.3 & 89.2 & 84.2 & 97.3 & 92.9 & -& 94.0& 88.6&  -&\textbf{0.99}\\
 HiSup \cite{xu2023hisup}& 79.4& 92.7& 85.3& 81.5& 93.1& 86.7& 66.5& 94.3& 89.6&  \textbf{0.726}&-\\ \midrule
 GeoFormer (ours) & \textbf{91.5} & \textbf{96.6} & \textbf{93.1} & \textbf{97.8} & \textbf{98.8} & \textbf{98.1} & \textbf{97.1} & \textbf{98.1} & \textbf{97.4} &  0.913 & \textbf{1.01} \\
 \bottomrule
\end{tabular}
}
\label{tab:main-results}
\end{table}

The quantitative results of our study are presented in Table \ref{tab:main-results}, while qualitative comparisons are illustrated in Figure \ref{fig:qual-results-main}. At first glance, Table \ref{tab:main-results} reveals the strong performance of GeoFormer across all metrics, exceeding prior state-of-the-art works by almost 12 percentage points in AP. A detailed examination of the qualitative results in Figure \ref{fig:qual-results-main} reveals nuanced differences in model performance. For instance, our model, GeoFormer, tends to slightly overestimate in certain instances (as seen in the lower right image), while the PolyWorld model shows a tendency to underestimate, particularly noticeable in the second row, middle image. Additionally, our model's high recall is evident in its tendency to occasionally identify extraneous objects as false positives, as demonstrated in the last row of Figure \ref{fig:qual-results-main}. The results are generated by the best trained model achieving the lowest validation loss, conducting conditional inference on the test-set of the Aicrowd dataset. We utilize Nucleus sampling with $p=0.95$ for all the results \cite{holtzman2019curious}.

Table \ref{tab:main-results} highlights the areas where GeoFormer excels, demonstrating superior performance in most evaluated metrics compared to competing methods. Despite trailing slightly behind methods like HiSup \cite{xu2023hisup} in PoLiS distance, GeoFormer exhibits a substantial lead in polygon fidelity and segmentation mask accuracy, achieving a remarkable 30\% improvement in bAP and a 7 percentage point increase in the polygon-specific complexity-aware IoU. 

GeoFormer's strong performance is attributed to its parameterised design, which effectively incorporates the sparse nature of building corner vertices, central to the prediction task without relying on multiple loss terms or complex post-processing steps. The intrinsic integration of the order and discrete properties of the vertices into the model's architecture and learning process enables GeoFormer to adeptly handle the challenging task of producing accurate multi-object vectorised representations of buildings from satellite imagery end-to-end. While GeoFormer may not lead in the PoLiS distance metric, it excels in delivering state-of-the-art performance across a majority of benchmark metrics and generates highly accurate polygonal representations, a key consideration in remote sensing and geospatial analysis.

%% file: sec/6_ablations_robustness.tex
\section{Ablations and robustness studies}

\begin{table}[tb]
\centering
\begin{minipage}{0.48\linewidth}
    \centering
    \caption{Inference results from our ablation studies trained on a subset of the Aicrowd dataset \cite{mohanty2020deep}. 
    }
    \label{tab:ablation-results}
    \vskip 0.05in
    \resizebox{\linewidth}{!}{%

    \begin{tabular}{cccccrr}
        \toprule
        \textbf{Sort polygons} & \textbf{Pyramid Features} & \textbf{ALiBi} & \textbf{RoPE} & \textbf{Mask} & \textbf{AP}$\uparrow$ & \textbf{bAP} $\uparrow$\\
        \midrule
        \checkmark & \checkmark & \checkmark & \checkmark & \checkmark & 12.89 & 31.07 \\
        \checkmark & \checkmark & \checkmark & \checkmark &  & \textbf{18.68} & \textbf{34.12} \\
        \checkmark & \checkmark & \checkmark &  &  & 12.10 & 36.73 \\
        \checkmark & \checkmark &  &  &  & 0.00 & 0.00 \\
        \checkmark &  &  &  &  & 0.16 & 0.13 \\
         &  &  &  &  & 0.00 & 0.00 \\
        \bottomrule
    \end{tabular}%

    }
\end{minipage}%
\hfill
\begin{minipage}{0.48\linewidth}
    \centering
    \caption{Rankings from four to one averaging across perturbation factors. Best performance highlighted in bold. 
    }
    \vskip 0.05in
    \resizebox{\linewidth}{!}{%

\begin{tabular}{lcccc}
        \toprule
        \textbf{Perturbation} & \textbf{FFL} & \textbf{PolyWorld} & \textbf{HiSup} & \textbf{GeoFormer} \\ \midrule
        Baseline & 4 & 3 & 2 &  \textbf{1} \\
        Downsample & 3 & 2 & \textbf{1} & 2 \\
        Dropout & 4 & 2 & 2 & \textbf{1} \\
        Rotation & 4 & 2 & 2 &  \textbf{1} \\ \bottomrule
    \end{tabular}%
    
    }
    \label{tab:robustness-average-rankings}
\end{minipage}
\vskip -0.1in
\end{table}

We conducted ablation studies to understand the role of the different embeddings and model properties, including pyramidal feature maps, polygon sorting, RoPE \cite{su2021roformer} embeddings, random token masking, and ALiBi localized attention heads \cite{press2021train}.
These experiments, along with robustness studies, were conducted on a smaller subset of the Aicrowd dataset, consisting of 8,366 training images with 71,871 building instances and 1,820 test images with 15,770 instances.
\paragraph{Model ablations}
Table \ref{tab:ablation-results} summarises the ablation studies. ALiBi and RoPE embeddings significantly impact model performance, highlighting their critical role. Polygon sorting and SWIN Pyramid Features also substantially affect the model's learning capability. While random masking appears to have a minor (or even potentially detrimental) effect, we find it contributes to more stable training and improved generalisation when fitting on the full training set. Further ablation studies are presented in the supplementary material.

\paragraph{Robustness experiments}%
We evaluated each method's robustness by exposing the best-trained models to various random perturbations commonly encountered in remote sensing, such as missing values (erased dropout \cite{zhong2020random} with 3, 6, 9, 12\% of pixels set to zero), reduced resolution (bilinear downsampling by factors 2, 3, 4, 5 from the original 30\,cm resolution to 60\,cm, 90\,cm, 1.2\,m, and 1.5\,m), and rotations (at $30^{\circ}$ increments, i.e., $30^{\circ}$, $60^{\circ}$, $90^{\circ}$, and $120^{\circ}$).
Table \ref{tab:robustness-average-rankings} shows the results of the robustness studies, and GeoFormer outperforms former incumbent methods in all but one scenario, sharing second place with PolyWorld in the downsampling case. We speculate that this discrepancy is due to the coarse approximation of the input image resolution using a $36\times 36$ feature map, which results in a more lossy spatial compression than the full image dimensions we seek to learn. Full results and evaluations of these robustness experiments are presented in the supplementary materials.

\subsection{Limitations of the auto-regressive tokenised approach}
Despite the challenges associated with conditional generative models, such as complexity and slower inference speeds compared to models like \cite{zorzi2022polyworld,girard2021polygonal,xu2023hisup,li2023joint-followup}, GeoFormer presents its advantages in the performance of this approach. However, this parameterization is not without limitations. For one, the necessity to perform multiple forward passes in order to extract all buildings in the image imposes computational constraints, which are clearly evident in Table \ref{tab:model_overview}. GeoFormer is composed of close to 25\% more trainable parameters and presents average inference speeds of 64 times slower than the fastest model HiSup \cite{xu2023hisup}, and similar methods with a rasterised approach. Additionally, the encoded image feature map in our study $I_{F}$ is $36 \times 36$, which requires the decoder to map the tokens from a lower dimensional representation into the larger image representation, leading to higher errors when the discrepancy between the input image resolution and the feature maps is too high. Finally, the GeoFormer works on one-dimensional token representations of each image dimension, while not a hindrance towards performance, we may save additional computation if modelling the problem with token pairs as opposed to the single dimensional tokens in our approach.

%% file: sec/7_discussion.tex
\section{Future works}

\begin{table}
\caption{Overview of parameter count and inference speed. Inference speeds are computed on a single Nvidia 3090 GPU.}
\vskip 0.05in
\centering
\resizebox{0.75\columnwidth}{!}{%
\begin{tabular}{llcc}
\toprule
\textbf{Method} & \textbf{Backbone} & \#\textbf{Params} & \textbf{Inference / image} \\ \midrule
Frame Field Learning  \cite{girard2021polygonal} & UResNet101 & $74.6 \mathrm{M}$ &  $0.047 \mathrm{~s}$ \\
PolyWorld \cite{zorzi2022polyworld} & R2U-Net & 39.0M & $0.078 \mathrm{~s}$ \\
HiSup \cite{xu2023hisup} & HRNetV2-W48 & 74.3M & $0.030 \mathrm{~s}$ \\ \hline
GeoFormer (ours) & SwinV2-Pyramid & 97.4M & 1.93s  \\\bottomrule
\end{tabular}
}
\label{tab:model_overview}
\vskip -0.15in
\end{table}%

The GeoFormer represents a significant advancement in building delineation using a single encoder-decoder architecture. Despite its performance improvements (Table \ref{tab:main-results}), it faces scalability challenges in larger scenes due to the quadratic complexity of multi-headed attention mechanisms \cite{vaswani2017attention}. Future research could explore alternative parameterisations for the likelihood function. While the categorical likelihood approach has been effective, a discretized mixture of logistics, similar to PixelCNN++ \cite{salimans2017pixelcnn++}, may be more optimal for two-dimensional geometric problems. This shift could potentially improve model training, compression, and inference/training speed, as the current approach requires precise matching of the target token at time \( t \) to decrease the negative log-likelihood. Additionally, it would be interesting to explore avenues for encoding positions in the embedded images to be of higher dimension such as 3D \cite{khomiakov2023learning} or perhaps even mapping one-to-one with the image resolution. We believe this would enable the model to learn better to associate the appropriate value to each of the decoded tokens. In addition, we are curious to explore a similar approach to solve wire-frame multi-polygon challenges, in which the same set of points needs to be re-visited. Such a problem may prove challenging for our model parameterization, and perhaps we can learn to associate each set of points with a particular frame of the wireframe.

%% file: sec/8_conclusion.tex
\section{Conclusion}
In this paper, we introduced GeoFormer, a novel image-to-sequence auto-regressive probabilistic model for predicting multiple object polygons. By combining feature pyramids \cite{lin2017feature} with the SWINv2 transformer backbone, GeoFormer represents a new approach to object delineation in satellite imagery, using a single long sequence with special tokens separating each object of interest. Our findings support the hypothesis that the sparse nature of polygon vertices is well-addressed by a parameterization closely aligned with the task's core objective: generating a compact and geometrically precise set of points forming closed polygons around buildings.

GeoFormer challenges conventional semantic segmentation methods and demonstrates the potential for generative models through state-of-the-art performance on the Aicrowd \cite{mohanty2020deep} satellite imagery benchmark. The model's ability to directly target the sparsity and precision of polygon vertices could lead to substantial advancements in this domain. Future research interests include exploring new parameterisations of the likelihood function and investigating multipolygon scenarios such as identifying linear surfaces of building roofs \cite{khomiakov2023learning}.

In conclusion, GeoFormer offers a meaningful contribution to object delineation in satellite imagery, outperforming prior works by a substantial margin and showing promise for further research and application in geospatial analysis using auto-regressive sequence-to-sequence models.

\section{Acknowledgements}
This work was supported by Otovo ASA through a collaborative research initiative. We thank them for their valuable contribution. We would also like to acknowledge the EuroHPC Joint Undertaking for awarding us access to Karolina at IT4Innovations, Czech Republic.

%% file: sec/X_suppl.tex
\section{Model setup}
The GeoFormer is composed of an image encoder and an auto-regressive decoder. The encoder uses a patch size of 4, a window size of 7, and SWINv2 encoder dimensions of 192. We apply a dropout rate of 0.2 in the SWINv2 layer paths, with depths of 2, 2, 18, and 2 for layers 1, 2, 3, and 4, respectively, each consisting of 6, 12, 12, and 48 attention heads. The encoder's hidden dimensions are 512. The decoder consists of 8 layers, each with 24 attention heads (8 dedicated to ALiBi attention), and dropout of 0.1 and 0.2 in the attention paths and feedforward layers, respectively. We use the Adam-W optimizer with a learning rate of $2\times10^{-4}$, $\beta = (0.9, 0.999)$, and a weight decay of $1 \times 10^{-2}$.

\section{Ablation studies}
\begin{table*}[ht]
\caption{Inference results from our ablation studies trained and validated on the small version of the Aicrowd dataset \cite{mohanty2020deep}}
\centering
\resizebox{\textwidth}{!}{
\begin{tabular}{cccccrrrrrrrrrrr}
\toprule
\textbf{Sort polygons} & \textbf{Pyramid Features} & \textbf{ALiBi} & \textbf{RoPE} & \textbf{Mask} & \textbf{AP}$\uparrow$ &  \textbf{AP}$_{50}$$\uparrow$ &  \textbf{AP}$_{75}$$\uparrow$ &    \textbf{AR}$\uparrow$ &   \textbf{AR}$_{50}$$\uparrow$ &   \textbf{AR}$_{75}$$\uparrow$ & \textbf{bAP}$\uparrow$ &    \textbf{IoU}$\uparrow$ &   \textbf{C-IoU}$\uparrow$  &  \textbf{N-ratio} \\
\midrule
\checkmark & \checkmark & \checkmark & \checkmark &  & 18.68 & 34.10 & 18.99 & 49.93 & 70.34 & 53.10 & 34.12 & 67.29 & 53.62 & 2.17 \\
 & \checkmark & \checkmark & \checkmark &  & 16.73 & 32.60 & 15.59 & 44.97 & 61.38 & 48.28 & 28.19 & 59.15 & 49.27 & 1.99 \\
\checkmark & \checkmark & \checkmark &  & \checkmark & 15.27 & 28.69 & 15.16 & 48.28 & 71.72 & 53.79 & 31.43 & 51.92 & 23.18 & 10.04 \\
\checkmark & \checkmark &  &  & \checkmark & 15.15 & 27.96 & 15.12 & 52.07 & 69.66 & 56.55 & 32.44 & 61.33 & 30.74 & 8.51 \\
\checkmark & \checkmark & \checkmark & \checkmark & \checkmark & 12.89 & 25.29 & 11.72 & 45.17 & 66.21 & 51.72 & 31.07 & 62.80 & 45.27 & 4.03 \\
 & \checkmark & \checkmark &  & \checkmark & 12.33 & 26.25 & 10.17 & 39.86 & 60.69 & 42.07 & 22.36 & 46.18 & 19.95 & 9.59 \\
\checkmark & \checkmark & \checkmark &  &  & 12.10 & 21.68 & 12.61 & 58.07 & 76.55 & 62.07 & 36.73 & 67.59 & 48.35 & 4.30 \\
\checkmark & \checkmark &  & \checkmark & \checkmark & 10.82 & 22.05 & 9.28 & 36.41 & 54.48 & 40.00 & 25.62 & 53.50 & 21.49 & 10.20 \\
\checkmark &  &  & \checkmark & \checkmark & 10.03 & 21.06 & 8.21 & 45.24 & 67.59 & 46.21 & 27.01 & 48.51 & 22.80 & 9.14 \\
 & \checkmark &  &  & \checkmark & 9.95 & 20.86 & 8.50 & 50.90 & 71.72 & 57.24 & 22.43 & 40.11 & 13.68 & 11.90 \\
 & \checkmark & \checkmark &  &  & 9.58 & 18.63 & 9.19 & 45.59 & 60.00 & 50.34 & 27.40 & 42.73 & 17.91 & 10.61 \\
 &  & \checkmark & \checkmark &  & 9.29 & 22.42 & 5.71 & 39.24 & 59.31 & 41.38 & 20.68 & 49.70 & 39.23 & 2.81 \\
\checkmark &  & \checkmark &  &  & 9.01 & 16.72 & 8.91 & 48.55 & 66.90 & 51.03 & 30.67 & 60.09 & 24.64 & 10.19 \\
\checkmark &  &  &  & \checkmark & 8.39 & 16.26 & 7.85 & 40.14 & 56.55 & 44.83 & 26.85 & 53.63 & 20.17 & 9.99 \\
\checkmark &  & \checkmark & \checkmark &  & 7.64 & 15.35 & 6.67 & 50.21 & 66.21 & 55.17 & 28.61 & 58.00 & 40.75 & 4.89 \\
 & \checkmark & \checkmark & \checkmark & \checkmark & 6.77 & 16.57 & 4.10 & 36.14 & 56.55 & 37.93 & 18.00 & 41.00 & 25.18 & 6.07 \\
 &  &  &  & \checkmark & 6.23 & 13.73 & 4.92 & 41.38 & 60.00 & 45.52 & 20.46 & 43.36 & 23.22 & 5.51 \\
\checkmark &  & \checkmark &  & \checkmark & 6.20 & 12.73 & 5.21 & 41.66 & 64.83 & 44.14 & 26.34 & 50.58 & 23.23 & 9.76 \\
\checkmark &  & \checkmark & \checkmark & \checkmark & 5.56 & 13.45 & 3.39 & 26.07 & 46.90 & 26.21 & 17.78 & 44.60 & 27.95 & 5.64 \\
 &  &  & \checkmark & \checkmark & 5.56 & 13.45 & 3.39 & 26.07 & 46.90 & 26.21 & 17.78 & 44.60 & 27.95 & 5.64 \\
 & \checkmark &  & \checkmark & \checkmark & 4.54 & 12.05 & 2.32 & 23.72 & 42.76 & 24.14 & 13.67 & 31.08 & 11.49 & 10.88 \\
 &  &  & \checkmark &  & 4.05 & 12.08 & 1.46 & 22.07 & 41.38 & 18.62 & 15.13 & 40.94 & 32.87 & 2.43 \\
 &  & \checkmark &  & \checkmark & 2.67 & 7.89 & 1.07 & 28.14 & 47.59 & 28.97 & 13.54 & 35.75 & 18.23 & 8.12 \\
 &  & \checkmark & \checkmark & \checkmark & 0.61 & 2.06 & 0.08 & 16.90 & 33.10 & 14.48 & 4.91 & 22.02 & 10.31 & 7.64 \\
\checkmark &  &  &  &  & 0.16 & 0.20 & 0.20 & 5.38 & 17.93 & 2.07 & 0.13 & 5.35 & 1.71 & 11.92 \\
 &  & \checkmark &  &  & 0.13 & 0.59 & 0.02 & 7.52 & 19.31 & 7.59 & 3.05 & 21.15 & 13.89 & 4.20 \\
 & \checkmark &  & \checkmark &  & 0.01 & 0.03 & 0.00 & 0.90 & 3.45 & 0.00 & 0.03 & 14.79 & 4.94 & 12.07 \\
\checkmark &  &  & \checkmark &  & 0.00 & 0.00 & 0.00 & 0.14 & 1.38 & 0.00 & 0.04 & 10.41 & 3.52 & 11.85 \\
 & \checkmark &  &  &  & 0.00 & 0.00 & 0.00 & 0.00 & 0.00 & 0.00 & 0.00 & 0.00 & 0.00 & 0.00 \\
\checkmark & \checkmark &  & \checkmark &  & 0.00 & 0.00 & 0.00 & 0.00 & 0.00 & 0.00 & 0.00 & 0.00 & 0.00 & 0.00 \\
\checkmark & \checkmark &  &  &  & 0.00 & 0.00 & 0.00 & 0.00 & 0.00 & 0.00 & 0.00 & 0.00 & 0.00 & 0.00 \\
 &  &  &  &  & 0.00 & 0.00 & 0.00 & 0.00 & 0.00 & 0.00 & 0.00 & 0.00 & 0.00 & 0.00  \\
\bottomrule
\end{tabular}
}
\label{tab:ablation-results_supplementary}
\end{table*}

We computed ablations on a combination of factors related to our model's contributions. These included applying the sorting of polygons in each image, incorporating an encoder with SWIN pyramidal feature maps \cite{liu2022swin}, the value of added linear bias (ALiBi) \cite{press2021train}, using relative rotational embeddings (RoPE) \cite{su2021roformer}, and adding random token masking during training in our decoder. The total number of ablation combinations for all experiments was 32. In the main paper, we highlighted 8 experiments, while in Table \ref{tab:ablation-results_supplementary}, we present the full results.

A primary observation was the significant limitation of our model in fitting the data in any form when applying our pyramidal features without including either ALiBi or mask-based training. In fact, simply adding random masked decoding, along with our pyramidal feature maps, improved performance by 10 percentage points in mAP. Although not evident in this chart, we observed that the model immediately overfitted the data if only the pyramidal feature maps were present, indicating that most learning occurred in the Encoder, leading to poor generalisation when performing auto-regressive decoding.

Another notable observation was that the absence of positional embeddings, particularly in the form of ALiBi, substantially decreased performance. Additionally, when using only either ALiBi or RoPE embeddings, the model became less capable of predicting the correct number of points in the final prediction (N-ratio).

Overall, we observed that performance was generally higher when introducing the pyramidal features, as opposed to not using them. The introduction of both ALiBi and RoPE embeddings, along with the sorting of polygons, allowed for the best performance in our model. While masked decoding provided somewhat lower performance than without it, we observed faster convergence and better performance with masking when training on the full dataset.

\subsection{Robustness studies}
\label{sec:robustness_studies}
\begin{figure}
    \centering
    \includegraphics[width=1\textwidth]{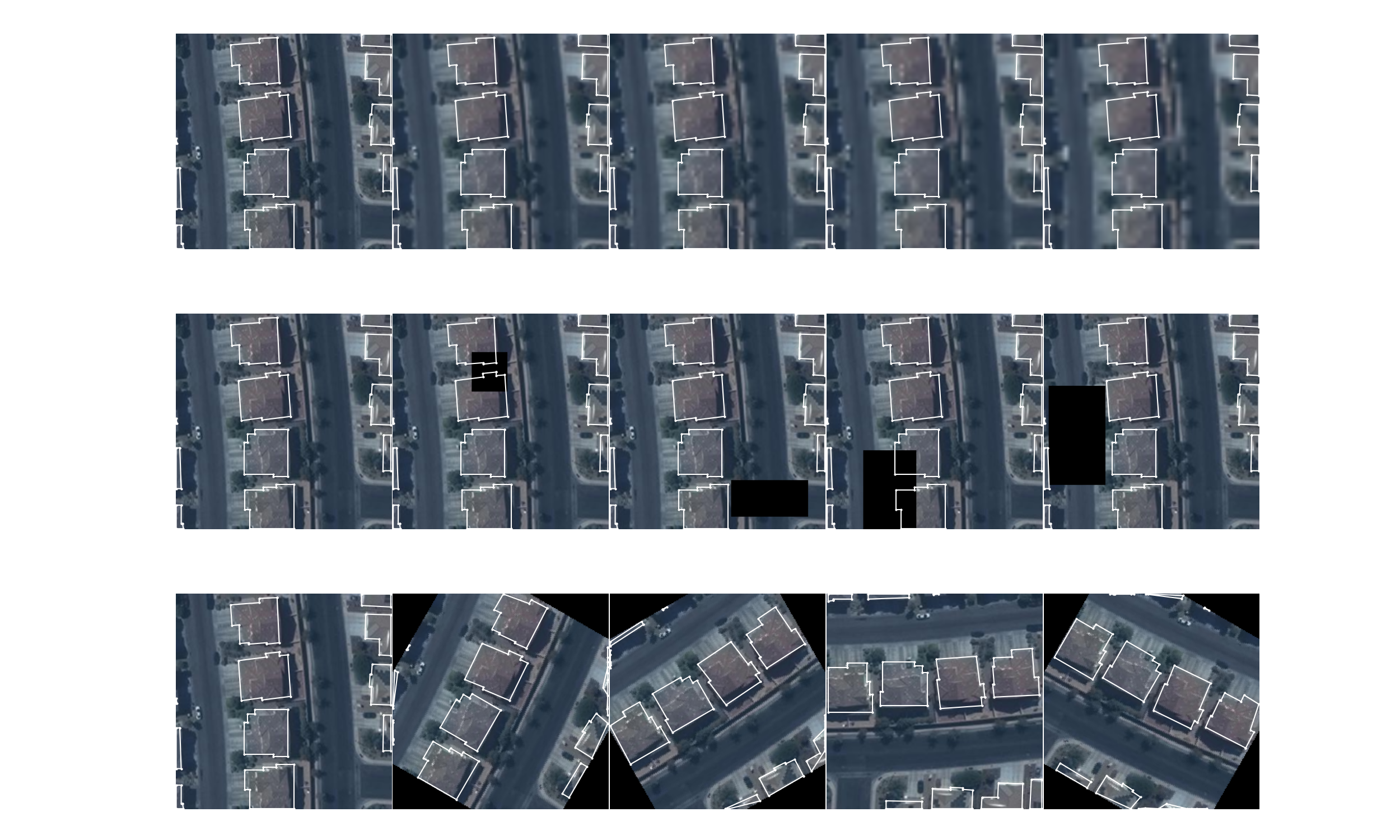}
    \caption{Visual examples of perturbations performed to input images in the robustness studies. From top row: downsampling, erased dropout, and rotations.}
    \label{fig:pertubation_visualizations}
\end{figure}

\begin{table*}[]
\caption{Robustness results on the small version of the Aicrowd dataset \cite{mohanty2020deep}}
\resizebox{0.98\textwidth}{!}{
\begin{tabular}{lrrrrrrrrrrrr}
\toprule
& \textbf{Model} &\textbf{AP}$\uparrow$ & \textbf{AP}$_{50}$$\uparrow$ & \textbf{AP}$_{75}$$\uparrow$ & \textbf{AR}$\uparrow$ & \textbf{AR}$_{50}$$\uparrow$ & \textbf{AR}$_{75}$$\uparrow$ & \textbf{bAP}$\uparrow$ & \textbf{C-IoU}$\uparrow$ & \textbf{IoU}$\uparrow$ & \textbf{N-ratio} &  \textbf{PF} \\
\midrule
\multirow{4}{*}{\rotatebox[origin=c]{90}{\textbf{Baseline}}} & GeoFormer & 90.8 & 95.75 & 91.94 & 99.38 & 100.0 & 99.31 & 97.03 & 97.32 & 98.02 & 1.0 & 0 \\
 & HiSup & 70.33 & 88.43 & 76.53 & 96.7 & 99.82 & 99.52 & 67.41 & 89.71 & 94.19 & 1.0 & 0 \\
 & PolyWorld & 52.65 & 80.07 & 58.16 & 76.34 & 83.52 & 80.32 & 55.01 & 79.57 & 83.87 & 0.91 & 0 \\
 & FFL & 37.44 & 61.6 & 40.93 & 82.94 & 97.57 & 92.18 & 49.86 & 27.85 & 81.19 & 5.91 & 0 \\
\midrule
\multirow{16}{*}{\rotatebox[origin=c]{90}{\textbf{Downsample}}} & HiSup & 62.1 & 83.95 & 69.67 & 92.73 & 99.61 & 98.95 & 61.75 & 86.54 & 91.48 & 1.0 & 2 \\
 & HiSup & 56.65 & 80.13 & 63.62 & 88.91 & 99.16 & 97.88 & 57.22 & 83.11 & 88.81 & 1.0 & 3 \\
 & HiSup & 46.13 & 70.1 & 51.98 & 79.83 & 96.71 & 91.15 & 48.2 & 75.12 & 82.56 & 0.98 & 4 \\
 & PolyWorld & 45.34 & 71.83 & 50.44 & 64.03 & 75.29 & 68.19 & 48.23 & 71.43 & 78.39 & 0.84 & 2 \\
 & PolyWorld & 33.42 & 57.93 & 36.33 & 45.63 & 58.81 & 50.34 & 37.47 & 56.64 & 66.65 & 0.74 & 3 \\
 & FFL & 32.84 & 56.44 & 35.33 & 78.41 & 95.42 & 88.68 & 45.94 & 25.82 & 78.22 & 6.29 & 2 \\
 & HiSup & 32.59 & 55.31 & 35.29 & 65.68 & 87.38 & 76.64 & 37.6 & 63.51 & 72.55 & 1.01 & 5 \\
 & GeoFormer & 31.4 & 56.03 & 31.93 & 59.86 & 74.48 & 64.83 & 64.48 & 57.82 & 67.4 & 0.78 & 2 \\
 & FFL & 24.3 & 45.58 & 24.13 & 73.1 & 94.07 & 81.13 & 40.38 & 21.72 & 73.2 & 7.32 & 3 \\
 & PolyWorld & 20.71 & 39.25 & 20.81 & 27.67 & 37.53 & 30.66 & 25.37 & 38.1 & 49.73 & 0.57 & 4 \\
 & GeoFormer & 17.0 & 35.25 & 14.83 & 48.62 & 63.45 & 51.03 & 57.6 & 39.43 & 49.84 & 0.68 & 3 \\
 & FFL & 12.13 & 27.23 & 9.51 & 54.99 & 77.63 & 60.38 & 26.93 & 16.07 & 60.27 & 9.29 & 4 \\
 & PolyWorld & 10.54 & 22.12 & 9.12 & 17.99 & 25.86 & 18.99 & 15.36 & 21.82 & 32.88 & 0.42 & 5 \\
 & GeoFormer & 5.25 & 12.23 & 3.91 & 35.66 & 51.72 & 37.24 & 53.48 & 16.83 & 26.27 & 0.38 & 4 \\
 & FFL & 3.68 & 9.62 & 2.2 & 43.02 & 67.92 & 46.09 & 14.43 & 11.63 & 47.0 & 11.29 & 5 \\
 & GeoFormer & 1.51 & 3.83 & 1.11 & 24.69 & 36.55 & 24.14 & 52.69 & 8.92 & 15.52 & 0.29 & 5 \\
\midrule
\multirow{16}{*}{\rotatebox[origin=c]{90}{\textbf{Dropout}}} & GeoFormer & 77.39 & 89.77 & 79.52 & 87.59 & 94.48 & 89.66 & 91.85 & 90.73 & 92.91 & 1.0 & 1 \\
 & GeoFormer & 66.04 & 82.09 & 69.5 & 79.52 & 93.79 & 81.38 & 87.26 & 84.27 & 87.73 & 0.98 & 2 \\
 & GeoFormer & 57.37 & 77.47 & 60.13 & 66.76 & 82.76 & 68.97 & 83.27 & 77.96 & 82.13 & 0.97 & 3 \\
 & HiSup & 56.97 & 75.73 & 61.13 & 83.65 & 93.18 & 85.29 & 60.91 & 77.05 & 82.62 & 1.11 & 1 \\
 & HiSup & 51.92 & 69.11 & 56.7 & 74.2 & 83.1 & 76.23 & 57.06 & 69.45 & 74.92 & 1.15 & 2 \\
 & GeoFormer & 50.49 & 72.34 & 53.17 & 59.66 & 83.45 & 58.62 & 80.52 & 74.38 & 79.31 & 0.93 & 4 \\
 & HiSup & 47.55 & 63.6 & 52.11 & 66.24 & 73.0 & 68.3 & 53.76 & 63.38 & 68.55 & 1.16 & 3 \\
 & PolyWorld & 46.95 & 74.81 & 50.46 & 69.36 & 81.46 & 73.68 & 51.0 & 73.77 & 79.33 & 0.88 & 1 \\
 & HiSup & 44.61 & 60.33 & 48.57 & 60.31 & 66.54 & 62.2 & 50.87 & 58.91 & 64.15 & 1.18 & 4 \\
 & PolyWorld & 44.2 & 71.32 & 47.4 & 61.14 & 75.97 & 64.07 & 48.94 & 70.08 & 76.43 & 0.85 & 2 \\
 & PolyWorld & 41.54 & 67.65 & 44.2 & 56.5 & 70.71 & 58.35 & 47.06 & 66.88 & 73.92 & 0.83 & 3 \\
 & PolyWorld & 39.59 & 65.33 & 41.98 & 47.8 & 63.39 & 49.43 & 45.1 & 63.08 & 70.73 & 0.8 & 4 \\
 & FFL & 33.12 & 56.57 & 35.0 & 77.63 & 95.69 & 87.87 & 47.78 & 26.41 & 78.4 & 6.32 & 1 \\
 & FFL & 30.39 & 53.21 & 31.68 & 71.37 & 94.61 & 79.25 & 46.0 & 25.5 & 75.79 & 6.56 & 2 \\
 & FFL & 27.89 & 49.64 & 28.7 & 64.31 & 89.49 & 66.85 & 44.33 & 24.79 & 72.87 & 6.85 & 3 \\
 & FFL & 26.27 & 47.08 & 26.65 & 58.01 & 84.64 & 60.65 & 43.09 & 24.3 & 70.58 & 6.9 & 4 \\
\midrule
\multirow{16}{*}{\rotatebox[origin=c]{90}{\textbf{Rotation}}} & GeoFormer & 70.52 & 93.32 & 80.72 & 98.48 & 99.31 & 99.31 & 83.62 & 91.98 & 92.64 & 1.0 & 3 \\
 & HiSup & 69.9 & 87.43 & 76.37 & 96.56 & 99.88 & 99.46 & 67.02 & 89.44 & 94.11 & 1.01 & 3 \\
 & PolyWorld & 51.33 & 78.99 & 57.49 & 74.16 & 81.84 & 79.54 & 54.09 & 78.1 & 82.94 & 0.9 & 3 \\
 & GeoFormer & 16.54 & 35.32 & 14.16 & 33.22 & 56.52 & 33.04 & 55.84 & 47.71 & 59.26 & 1.34 & 2 \\
 & GeoFormer & 15.82 & 33.47 & 13.36 & 35.26 & 56.14 & 42.11 & 55.82 & 46.91 & 58.28 & 1.36 & 1 \\
 & GeoFormer & 14.87 & 32.88 & 11.48 & 34.74 & 57.02 & 37.72 & 54.15 & 46.04 & 57.28 & 1.19 & 4 \\
 & PolyWorld & 14.54 & 29.47 & 13.61 & 19.84 & 33.24 & 21.8 & 18.12 & 30.37 & 45.86 & 0.42 & 1 \\
 & HiSup & 14.33 & 26.05 & 15.24 & 59.14 & 85.66 & 68.57 & 26.79 & 41.65 & 48.25 & 1.53 & 1 \\
 & PolyWorld & 14.3 & 29.18 & 13.35 & 19.84 & 31.88 & 22.34 & 18.18 & 30.65 & 46.17 & 0.44 & 4 \\
 & HiSup & 14.08 & 25.85 & 14.8 & 58.62 & 84.95 & 67.49 & 26.62 & 41.64 & 48.24 & 1.45 & 4 \\
 & HiSup & 14.05 & 25.92 & 14.8 & 58.51 & 84.82 & 67.93 & 26.6 & 41.89 & 48.78 & 1.53 & 2 \\
 & PolyWorld & 13.13 & 27.26 & 11.78 & 24.36 & 40.0 & 26.67 & 17.5 & 30.33 & 46.42 & 0.43 & 2 \\
 & FFL & 0.0 & 0.0 & 0.0 & 0.0 & 0.0 & 0.0 & 0.01 & 0.54 & 0.94 & 1.21 & 1 \\
 & FFL & 0.0 & 0.0 & 0.0 & 0.0 & 0.0 & 0.0 & 0.01 & 0.55 & 0.95 & 1.2 & 2 \\
 & FFL & 0.0 & 0.0 & 0.0 & 0.0 & 0.0 & 0.0 & 0.01 & 0.56 & 1.04 & 1.19 & 3 \\
 & FFL & 0.0 & 0.0 & 0.0 & 0.0 & 0.0 & 0.0 & 0.01 & 0.54 & 0.92 & 1.19 & 4 \\
\bottomrule
\end{tabular}
}
\label{tab:robustness-results_supplementary}
\end{table*}

\begin{figure*}
    \centering
    \includegraphics[width=\textwidth]{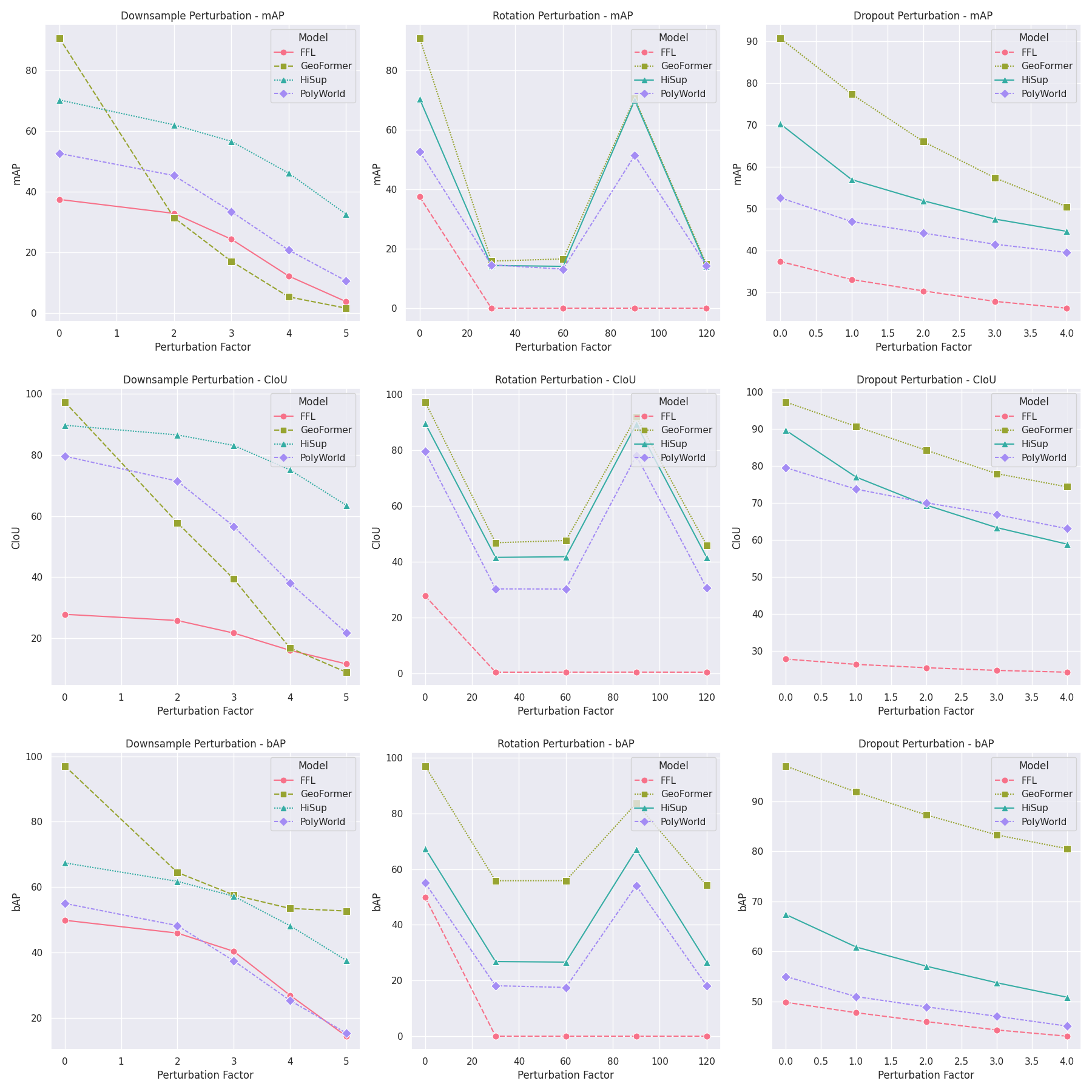}
    \caption{Performance relative to perturbations performed on the Aicrowd small dataset. We perform downsampling, rotations and random dropout. For downsampling a perturbation factor of 2 would equate to a 2x lower spatial resolution, while for dropout, each perturbation factor corresponds to $3\% \times$perturbation factor of pixels that are erased, while for the rotations the perturbation factor is the angles by which the input is rotated.}
    \label{fig:supplementary-robustness}
\end{figure*}

To better understand the situations where GeoFormer outperforms previous methods, we conducted a series of robustness studies. These studies were designed to simulate artefacts typically encountered in remote sensing imagery, including variations in spatial image resolution, rotational changes, and missing values. A visual example of the perturbations performed is illustrated in Figure \ref{fig:pertubation_visualizations}. The results are presented in Table \ref{tab:robustness-results_supplementary} and Figures \ref{fig:supplementary-robustness}. As mentioned in the main paper, we used the smaller Aicrowd dataset \cite{mohanty2020deep} for computing metrics, while employing the best model checkpoints trained on the full dataset. In Table \ref{tab:robustness-results_supplementary}, we present metrics similar to those in the main paper, with additional columns for Perturbation and Perturbation Factor (PF). Initially, we computed a baseline for each model, representing the performance across the chosen metrics without any perturbations. We followed the same approach as in the main paper, where COCO-metrics are computed as defined in the original MS-COCO benchmark \cite{lin2014microsoft}, while C-IoU and boundary average precision (bAP)\cite{cheng2021boundary} are calculated between the predicted polygon that overlaps the most with the ground truth (requiring a minimum of 0.5 overlap).

Our results show interesting trends across various perturbations. FFL \cite{girard2021polygonal} seems to rapidly deteriorate in performance regardless of the perturbation type. Meanwhile, both PolyWorld \cite{zorzi2022polyworld} and GeoFormer (our model) exhibit significant deterioration upon image downsampling, as seen in Table \ref{tab:robustness-results_supplementary} and Figure \ref{fig:supplementary-robustness}. In contrast, HiSup \cite{xu2023hisup} also shows degradation in the downsampling scenario, but to a lesser extent. For rotation and dropout perturbations, GeoFormer demonstrates stronger robustness compared to competing methods HiSup, PolyWorld, and FFL. We also observe that GeoFormer is quite robust in terms of boundary average precision \cite{cheng2021boundary} in all scenarios except image downsampling, as illustrated in Figure \ref{fig:supplementary-robustness} and Table \ref{tab:robustness-results_supplementary}. Overall, GeoFormer is on par with or slightly better than competing methods in handling image rotations, while it falls behind in scenarios involving image downsampling. However, it excels relative to other methods in dealing with missing values. The latter is likely due to masked training, while the robustness to rotations is attributed to rotation augmentations during training. The reason for its underperformance in downsampling scenarios is likely explained by the small feature map of 36x36 it needs to decode tokens from, but warrants further investigation.  

\subsection{Visualising attention maps}

We can visualise how the attention mechanism weights the image for each token \(s_{t}\). This visualisation involves projecting the normalised attention scores onto the input image, following a bi-linear upsampling to match the image's dimensions. We calculate the normalised scores from \(\hat{z}\) by averaging across all \(M\) attention heads, as follows:

\begin{align}
    \mathbf{Q}&= xW^{q}, \mathbf{K} = xW^{k} \\
    \hat{z}^{L} &= \frac{1}{M}  \sum_{m=1}^{M}  \sigma\left( \frac{\mathbf{Q}^{(m)}\mathbf{K}^{(m)^T}}{\sqrt{d}} \right) 
\end{align}

Here, \(\mathbf{Q}\) is derived from the decoder tokens, and the keys \(\mathbf{K}\) are represented as \(I_{F} \in \mathbb{R}^{36 \times 36 \times C}\), which is the image feature map comprising \(36 \times 36\) image patches with \(C\) hidden dimensions. \(\hat{z}^L\) represents the averaged attention scores from the final layer of the decoder.

\begin{figure*}
    \centering
    \includegraphics[width=1\textwidth,height=0.9\textheight]{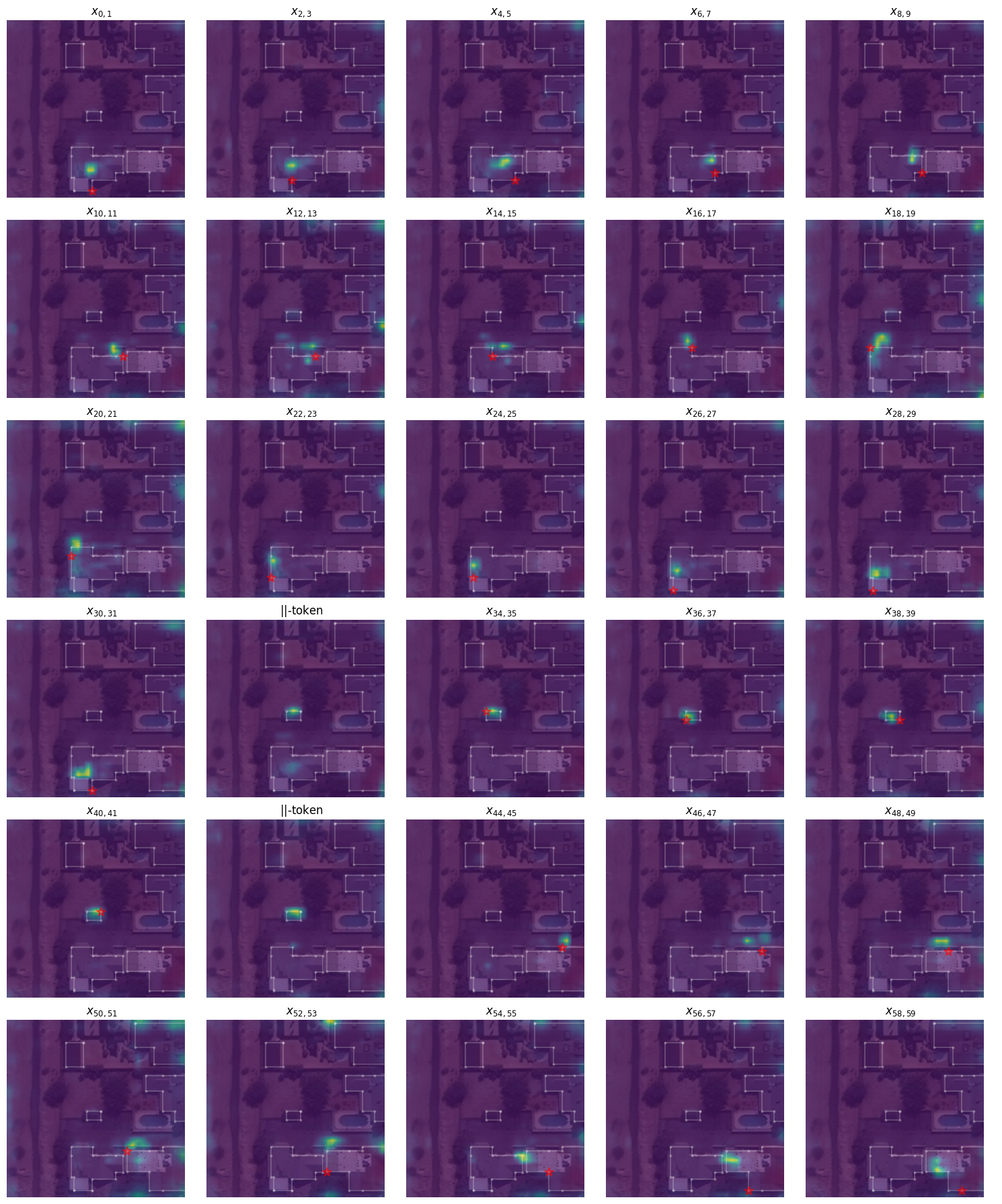}
    \caption{Visualisation of the attention maps on top of the input image and predicted polygons for pairs of tokens \(s_{t:t+1}\) from the final layer of the GeoFormer decoder.}
    \label{fig:attention_visualization}
\end{figure*}

The visualisation is shown in Figure \ref{fig:attention_visualization}, where we display the attention maps for inferred samples in token pairs. This means averaging over two consecutive token pairs, resulting in each image being composed of an \(x,y\)-pair. The images include fully predicted polygons in white, the input RGB image overlaid with attention scores, and a red star indicating the predicted coordinate at the paired timestep \(s_{t:t+1}\). We observe how the model shifts its attention towards the area of the building object it aims to predict and how the attention shifts upon the completion of the object, denoted by the special separator \(||\)-token.

\clearpage